\documentclass[twoside]{article}

\long\def\cut#1{}

\cut{TODOs for after paper submission:

add in reference to NEURAL PROGRAM SYNTHESIS WITH PRIORITY QUEUE TRAINING

Use top 5 in trianing set to initialize greedy searches--not random

add in Q=0.99 to sensitivity plots in SI (because was used in paper)
----------------------------

AM without the GAN prior, 

run Bombarelli

-----------------------------

compare to standard combinatorial optimizer (or ILS as per Roded Sharan) in noiseless case, or assuming it is noiseless even if it is not

VAE Dead units? Do we have any? can we comment on this?

sergey's comments?

Should what units protein expression is in to figure captions, and elsewhere if not already there.

Replace "sequence/expression" with "sequence-expression".

Take a look at references I forwarded from Tamara Broderick. I would add these in---she knows what she is talking about (and was a student with Mike J).

Add in our email addresses to the title sequence. Usually people would do it like {david, jennl}@berkely.edu.

mention that Killoran aren't actually using p(z) as a prior, although it is still effectively having an effect.

cite other related works, e.g. Gujimaraes et al 2018, etc.

explain how to do for classification instead of regression oracle

is Killoran sensitive to initilaization?

include times---Ron Shamir wanted to know this

pseuod code for multi-class classification instead of regression, and also for specification, in SI.

}

\usepackage{xr}
\usepackage{amsmath}
\usepackage{bm}
\usepackage{graphicx}
\usepackage{subcaption}
\usepackage{floatrow}
\usepackage[font=footnotesize,figurewithin=none]{caption}
\usepackage{amssymb}
\usepackage{bbm}
\usepackage{cite}
\usepackage{todonotes}
\usepackage{algorithm}
\usepackage{textcomp}
\usepackage[noend]{algpseudocode}

\newcommand{\eg}{\textit{e.\,g., }}
\newcommand{\ie}{\textit{i.\,e., }}
\newcommand{\etal}{\textit{et al}.}

\newcommand{\boldtheta}{\boldsymbol{\theta}}
\newcommand{\boldthetahat}{\hat{\boldsymbol{\theta}}}
\newcommand{\boldthetatplus}{{\boldsymbol{\theta}}^{(t+1)}}
\newcommand{\boldthetat}{\boldsymbol{\theta}^{(t)}}
\newcommand{\xbf}{\textbf{x}}
\newcommand{\xbfhat}{\hat{\textbf{x}}}

\newcommand{\textapprox}{\raisebox{0.5ex}{\texttildelow}}

\newcommand{\ourmethod}{DbAS}

\newcommand{\initset}{\{\xbf_\text{init}\}}

\usepackage{comment}

\usepackage{aistats2019}

\DeclareMathOperator*{\argmax}{argmax}

\begin{document}

%

%

\twocolumn[
\aistatstitle{Design by adaptive sampling}
\aistatsauthor{ David H. Brookes \And Jennifer Listgarten}
\aistatsaddress{University of California, Berkeley \And  University of California, Berkeley} 
]

\begin{abstract}

We present a probabilistic modeling framework and adaptive sampling algorithm wherein unsupervised generative models are combined with black box predictive models to tackle the problem of {\it input design}. In input design, one is given one or more stochastic ``oracle" predictive functions, each of which maps from the input design space (\eg DNA sequences or images) to a distribution over a property of interest (\eg protein fluorescence or image content). Given such stochastic oracles, the problem is to find an input that is expected to maximize one or more properties, or to achieve a specified value of one or more properties, or any combination thereof. We demonstrate experimentally that our approach substantially outperforms other recently presented methods for tackling a specific version of this problem, namely, maximization when the oracle is assumed to be deterministic and unbiased. We also demonstrate that our method can tackle more general versions of the problem.

\end{abstract}

\section{INTRODUCTION}


A convergence of recent advances in machine learning has set the stage for a solution to an important class of problems wherein one is given an arbitrary function and would like to find the setting(s) of the inputs that satisfy user-defined criteria on the function values. For example, given a fully-trained predictive model such as a neural network, one may seek to find an input setting that maximizes a particular class probability, such as the probability that an image is of a dog. This problem bears similarity to that tackled by activation-maximization (AM)~\cite{Simonyan2014, NguyenDYBC16}, which is typically used to visualize what a neural network has learned. However, it turns out that there are many application domains where one would like to solve a related technical problem---to design an object, represented by the input of a predictive model, that  yields particular predicted outputs. We call this {\it input design}. For example, one may want to design DNA sequences that maximize the amount of protein expression~\cite{Gustafsson2012}, or degree of protein binding~\cite{Killoran2017}; or design protein (amino acid) sequences that maximize properties such as secondary structure~\cite{Gupta2018}, or fluorescence\cite{Sarkisyan2016}. In fact, this general design problem is one of the fundamental unsolved problems in bioengineering. Currently, a painstaking, iterative, local search is performed in the laboratory, a procedure called {\it directed evolution}~\cite{Chen1991}. Herein, we propose a method for {\it in silico} directed evolution which can move through the design space more efficiently. However, our method is general and also has applications in other domains such as image design, circuit design, and molecule design~\cite{Gomez-Bombarelli2018}. Although AM is one way to tackle our design problem, as we demonstrate herein, another altogether different approach is better suited to the problem.

Beyond designing inputs to maximize a property, one may want to design inputs to simultaneously maximize several properties~\cite{Killoran2017}; to satisfy one property and maximize another; or to perform {\it specification} rather than maximization. In the specification problem one wants to design an input so as to achieve a particular property value or set of values, rather than to maximize it. For example, one may want to design a protein to fluoresce at a particular wavelength. 

For our design problems we assume that we have access to an ``oracle" predictive model, which, given an input such as a DNA sequence, returns a distribution over the properties of interest\cut{, or samples from that distribution}. Our design approach will specifically leverage the uncertainty in this distribution although it can also work with deterministic predictions. The oracle could be a neural network or Gaussian Process (GP) regression model~\cite{rasmussen2004} trained on data that we don't have access to; it could have been hand-created by domain experts; it could consist of real, physical measurements such those obtained in a laboratory, or even human-generated labels from an on-line crowd-sourcing platform. Additionally, the oracle need not be differentiable. In other words, the oracle is a black box that need only provide an input to output mapping.

Recently, several solutions to the maximization version of the input design problem (as opposed to specification) have been proposed~\cite{Gomez-Bombarelli2018, Killoran2017, Gupta2018}, which we now discuss. 

Gomez-Bombarelli \etal~design chemical structures represented as strings~\cite{Gomez-Bombarelli2018}. The approach taken is to (1) learn a neural-network-supervised variational auto-encoder (VAE) latent space so as to order the latent space by the property of interest, (2) build a GP regression model from the latent space to the supervised property labels, (3) perform gradient-based optimization over the latent space to maximize the GP function, (4) decode the optimal solution point(s) using the VAE decoder. The main drawbacks of this method are that it requires newly training supervised oracle-like models as part of the approach (from latent space to property space), as well as access to data needed to do so. Specifically, two oracle-like models must be trained: one to help shape the latent space, and another to do property maximization. Each model requires either feature/kernel engineering, architecture selection, or both. Moreover, the success of the overall method is highly dependent on these choices because these supervised models map from the latent space to the property space, thereby interacting with the entire approach. For these reasons, we did not to compare to this method.
\cut{Additionally, this method also employs a fixed prior that is dependent on the chosen training set, and will this suffer from the inherent difficulties discussed above concerning fixed priors in design scenarios. Finally, in many domains, excellent predictive models already exist which we would like to use directly out of the box, which would not be possible here.}

Killoran \etal~\cite{Killoran2017} use AM with a generative-adversarial network (GAN)~\cite{Goodfelow2014} prior on the input space~\cite{NguyenDYBC16} for a DNA sequence design problem. This approach has several limitations, many of which were already mentioned in ~\cite{Gupta2018}. First, it requires that the predictive model of the desired properties be differentiable,\footnote{And with non-vanishing gradient.} whereas in many domains, one may have well-established expert models that are not. Moreover, the data needed to create a differentiable version of the model may not be available. 

\cut{However, an even greater potential problem with~\cite{Killoran2017} is that the particular prior that is used influences the results of the optimization in an unpredictable way. In particular, Killoran \etal~train a GAN on one fixed training data set and never re-train it. This GAN acts as a prior for the AM, presumably to enforce realism-type constraints. This prior may assign little or no probability to near-optimal points of the input space, in which case any optimization heavily influenced by the prior will be unlikely to find the desired inputs}

\cut{However, it is not possible to understand or control the impact of the prior in this setting where it is critical to do so. One could only try different priors arising from training the GAN on different data sets or architectures and then proceed with one, fixed prior. Unlike in classical exponential-family Bayesian statistics, there is no notion of pseudo-counts or certainty of the prior to help understand what prior may be suitable. Moreover, there is in general no way to check the impact of the choice of prior by way of computational experiments since the purpose of the prior is to capture regularities that we may not know about (with images the visual observer can check, but this is not generalizable to other domains).}

\begin{figure*}[t]
    \centering
    \begin{subfigure}[t]{0.45\textwidth}
        \centering
        \includegraphics[width=\textwidth]{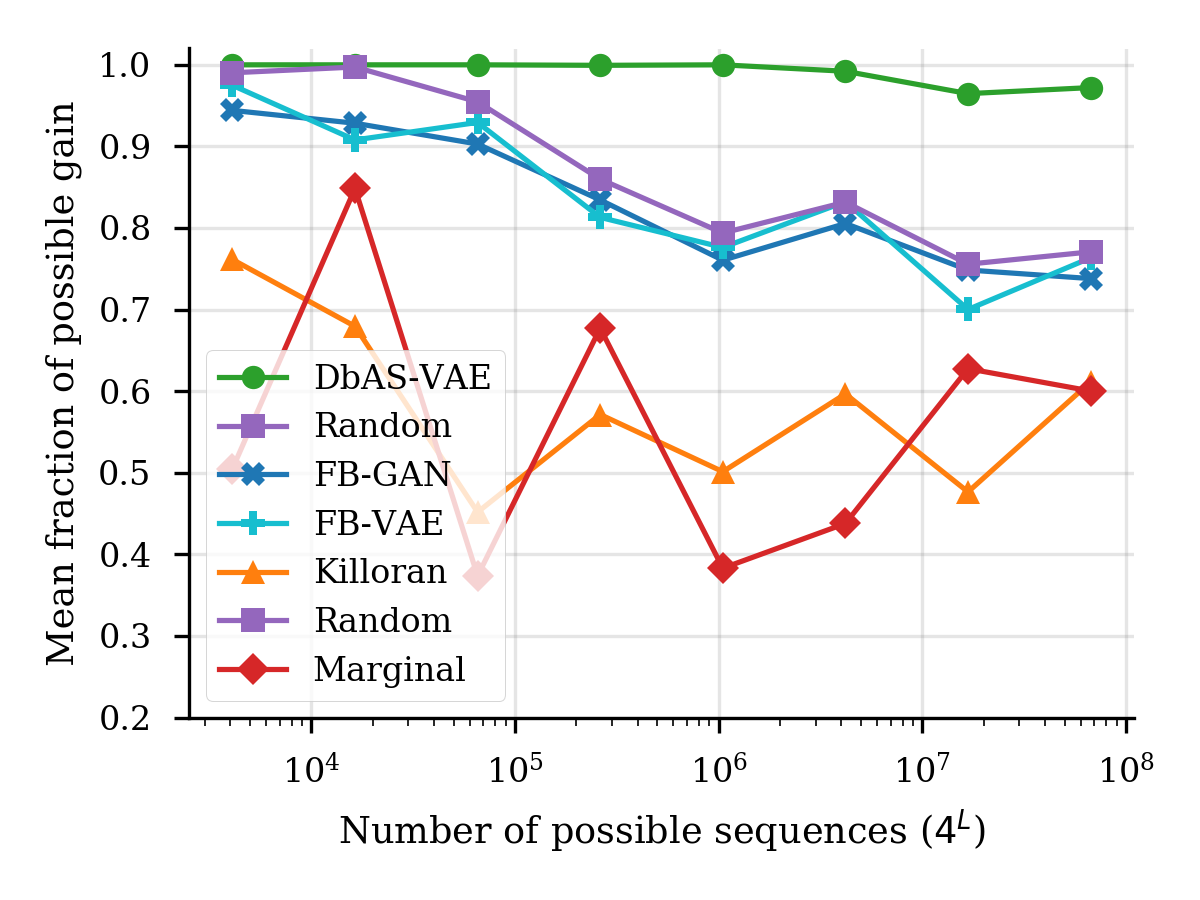}
        \caption{}
    \end{subfigure}
    \begin{subfigure}[t]{0.45\textwidth}
        \centering
        \includegraphics[width=\textwidth]{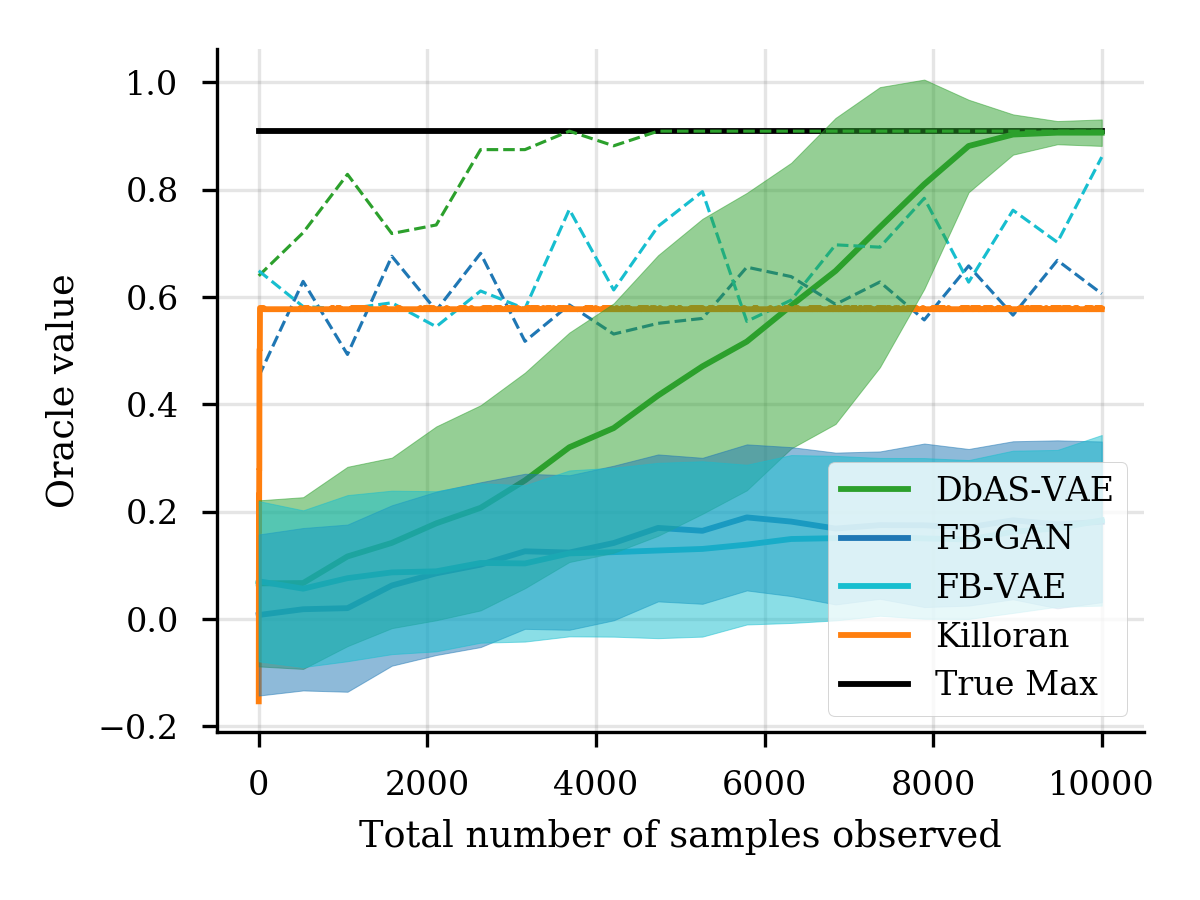}
        \caption{}
    \end{subfigure}
    \caption{Maximization of random, noise-free oracles. (a) Mean values of the {\it fraction of possible gain} (Section \ref{random_nn_section}) across 10 runs, for $L$ from 6 to 13. The horizontal axis shows the size of the search space at each $L$ (b) Representative trajectories for iterative methods at $L=8$. Solid lines, shaded regions and dashed lines correspond respectively to the mean, $\pm$ one standard deviation, and maximum values of oracle evaluations on sequences sampled at an iteration.}
    \label{fig:figure1}
\end{figure*}

\cut{In the Supplementary Information we show that this approximation can have extremely dire consequences. \todo{ADD to SI or remove sentence} As we shall see, the discreteness problem can be completely side-stepped with a different kind of approach altogether. Finally, as we will demonstrate, our newly-developed approach performs substantially better experimentally (Figure \ref{fig:figure1}). }

Gupta \etal~combine a GAN with a predictive model~\cite{Gupta2018}, the ``analyzer", much like our oracle, which need not be differentiable. The method proposed is to (1) generate samples from a GAN for the first iteration (trained on an initial set of objects such as DNA sequences), (2) use a pre-trained property model to predict the property of each generated sample such as whether or not a protein is antimicrobial, (3) update the sample set by replacing the oldest $n$ samples with the $n$ samples from step 2 that exceed a user-specified threshold that remains fixed throughout the algorithm (and where $n$ at each iteration is determined by this threshold), (4) retrain the GAN on the updated set of samples from step 3 for one epoch, (5) repeat. A stopping criterion is not provided. The intended goal is that as iterations proceed, the GAN will tend to produce samples which better maximize the property. Their procedure does not arise from any particular formalism, rather, it is described. Finally, unlike our method, this procedure does not handle uncertainty in quantitative property analyzers/oracles.

One issue with all three of these methods is the implicit (or explicit) use of a prior on the design space, which prevents the approach from appropriately exploring. For example, in ~\cite{Gomez-Bombarelli2018}, the design space is constrained by the fixed VAE, trained on some initial training data set. In the case where, for example, one is performing maximization, this may unnecessarily restrict the design space. Similarly, in ~\cite{Killoran2017}, the design space is constrained by a fixed GAN, trained on some initial data set. Finally, in ~\cite{Gupta2018}, although the method can in principle ``walk away" from the initially trained GAN, their procedure does not tend to do this because of the ``old" samples that tend to be retained. In fact, they cite this retention as a benefit of their approach, explaining that it helps to constrain search to feasible regions of the design space. We hypothesize, however, that it is for this reason that our method performs better. As shall become apparent, our procedure does not have this problem, yet can still enforce any necessary prior information by way of additional oracles, such as is done in Section \ref{expmax}.

\cut{While priors can be useful for Bayesian modelling and may be necessary for certain classes of problems, we demonstrate in this work that use of a prior can overly constrain the allowable space of an input design problem. \cut{That is, unless that allowable space is both well-understood and can be clearly specified in the prior}. A further concern about the approach by Killoran \etal~is with its ability to handle discrete input spaces. Because with AM the gradient needs to be propagated back through the latent space, such an approach cannot be directly applied to generative models that output discrete data, as the gradient would in general be zero. Killoran \etal~side-step this issue by having the GAN output a latent variable representing the soft-max probabilities of the discrete sequence, similar to the Gumbel-softmax approach for approximately reparameterizing distributions of discrete random variables~\cite{Jang2017}, which is differentiable. However, they do not anneal this approximation towards to the true discrete sample space. }

Our approach, {\it Design by Adaptive Sampling} (\ourmethod), is similar in spirit to the approach of Gupta \etal, but with a few key differences that place our method on a more rigorous foundation and yield substantial performance benefits (Figure \ref{fig:figure1}). Intuitively, the key differences have to do with (1) replacing the {\it ad hoc} sample selection scheme with one that emerges from a principled framework that we develop based on probabilistic modeling and sampling, (2) leveraging the results of that framework to obtain an appropriate sample selection and sample weighting scheme. 
\cut{Additionally, because we anneal to the desired properties, rather choose a fixed, user-specified threshold, we end up with a generative model which is more likely to yield the desired properties.}
Finally, we demonstrate how to use our approach for specification as well as maximization. Our approach is related to several others which are discussed after presentation of our method.
\cut{In our detailed methods, we describe how our approach is related to the Cross Entropy Method~\cite{Rubinstein1999,Rubinstein1997}, Covariance Matrix Adaptation Evolution Strategy (CMA-ES) \cite{CMA-ES} and others.}

It is worth noting that the problem of input design in discrete sequence space is related to the problem of combinatorial optimization. However, in combinatorial optimization problems,  the function being optimized is typically not stochastic, whereas in most real-world input design problems, these functions are stochastic approximations of a ground truth that is unknown; the ability to leverage this uncertainty, especially when the noise is heteroscedastic, will be extremely important.

Next we go over the derivation and details of our method before comparing it with other approaches experimentally.

\section{METHODS}

\paragraph{Preamble}

Our problem can be described as follows. We seek to find the setting(s) of the $D$-dimensional random vector, $X$ (\eg representative of DNA sequences or images), that satisfies some property desideratum. For example, we may want to design a protein that is maximally fluorescent (the {\it maximization} problem), or that emits light at a specific wavelength (the {\it specification} problem). In our running examples, we will focus on the case where $X$ is discrete, with realizations, $\xbf \in \mathbb Z^L$, because we are particularly interested in problems of sequence design. However, our method is immediately applicable to $X \in \mathbb{R}^L$.

We assume that we are given a scalar property predictor ``oracle", $p(y|\xbf)$, which provides a distribution over the property random variable, $Y \in \mathbb{R}$, given a particular input $\xbf$. From this oracle model we will want to compute the probability of various sets, $S$, occurring. For maximization, $S$ will be the set of values $y$ such that $y \geq y_\text{max}$ (where \mbox{$y_\text{max} = \max_\textbf{x} \mathbb{E}_{p(y|\textbf{x})}[y]$}). 
In the development of our method, we will also want to consider sets that correspond to less stringent criteria, such as $S$ corresponding to the set of all $y$ for which \mbox{$y \geq \gamma$, with $\gamma \leq y_\text{max}$}.
In specification, $S$ will be the event that the property takes on a particular value, $y=y_\textit{target}$ (strictly speaking, an infinitesimal range around it).
From our oracle model, we can calculate the conditional probability of these sets---that is, the probability that our property desideratum is satisfied for a given input---as \mbox{$P(S|\textbf{x}) \equiv P(Y\in S|\textbf{x}) = \int p(y|\textbf{x}) \mathbbm{1}_{S}(y) \, d y$} (where $\mathbbm{1}_{S}(y)=1$ when $y \in S$ and 0 otherwise). For the case of thresholding a property value, this turns into a cumulative density evaluation, \cut{\int_{y=\gamma}^{y=\infty} p(y|\textbf{x}) \, d y = }
\mbox{$P(S|\textbf{x}) = p(y \geq \gamma|\textbf{x}) = 1 - CDF(\xbf,\gamma)$}. As we shall see, it is by way of these cumulative density function (CDF) calculations that our method will appropriately leverage the uncertainty of the oracle.

If the oracle is not probabilistic, we can simply assume that it is so, with zero-noise. The extent to which this oracle is correct in expectation and variance, and confident, will determine the efficiency and accuracy of our method. Herein we assume an unbiased oracle such that in expectation, the predictions it makes are correct. We first outline our approach in the case where our aim is to perform maximization of a single property\cut{when there is only one property to consider, and we perform maximization}. Details of how to readily generalize this to the specification problem, and to more than one property, including a mix of maximization and specification, are in the Supplementary Information.


\cut{We assume that we have access to a scalar property predictor, or ``oracle", $p(y|\xbf)$, which provides a density over realizations, $y \in \mathbb{R}$ of a property random variable, $Y$, given a particular $\textbf{x}$. In our development, we consider a continuous property, but our method is trivially extendable to discrete $Y$. Now let $S$ be a set of desired values of the property $Y$. }

\cut{For example, in a specification context, $S$ is the set of values $y$ that lie in an infinitesimal range around a specified value, $y_0$, \ie~$S=[y_0 -\epsilon, y_0+\epsilon]$ for a target value $y_0$ and small $\epsilon >0$. Similarly, in a maximization context, $S$ could be the set of values greater than a small deviation around a maximum expected value, i.e. $S = [y_\text{max} - \delta y, \infty)$ where $y_{\text{max}} =  \max_\textbf{x} E_{p(y|\textbf{x})}[y]$. }

\cut{We can also consider target sets corresponding to significantly less restrictive criteria, such as the set of all $y$ that are greater than a certain threshold, \ie for which $y \geq y_0$. From our oracle model, we can calculate the conditional probability of these sets, as $P(S|\textbf{x}) \equiv P(Y\in S|\textbf{x}) = \int p(y|\textbf{x}) \mathbbm{1}_{S}(y) \, d y$. For the thresholding and maximization cases mentioned above, this is simply a cumulative density evaluation, \cut{\int_{y=y_0}^{y=\infty} p(y|\textbf{x}) \, d y = }
\mbox{$P(S|\textbf{x}) = P(y \geq y_0|\textbf{x}) = 1 - F_{Y|X=\textbf{x}}(y_0)$}. As we shall see, it is by way of these $P(S|\xbf)$ calculations that our method will appropriately leverage the uncertainty of the oracle. If the oracle is not inherently probabilistic, we can simply assume that it is so, with zero-noise,in which case, $P(S|\textbf{x})$ collapse to indicator function of the set $S$. The extent to which the oracle is correct and confident will determine the extent to which our approach, or any other, can work. Herein we assume an unbiased oracle such that in expectation, the predictions it makes are correct. We first outline our approach when there is only one property to consider, and later extend it to multiple properties.
}

\paragraph{Our approach}

We cast the problem of input design as one of maximizing the probability of satisfying the desired property. Recall that $S$ is the set of property values that satisfy our desideratum. If we could maximize the expected probability that our desideratum is satisfied, where expectation is performed over a generative model distribution, $p(\textbf{x}|\boldtheta)$, given by say, a VAE, then we could sample from the generative model to obtain our solution. In particular, if (a) we could solve the following optimization problem,
\begin{align}
\boldthetahat &= \argmax_{\boldtheta} \log P(S|\boldtheta)
\\
&= \argmax_{\boldtheta} \log \int p(S|\xbf) p(\xbf|\boldtheta) \, d\xbf
\\
&= \argmax_{\boldtheta} \log \mathbb{E}_{p(\xbf|\boldtheta)} \left[ P(S| \xbf) \right],
\label{eq:marginal}
\end{align}
(b) the class of generative models was of sufficiently high capacity (c) $S$ was an infinitesimally small set, and (d) the oracle was noise-free,
then $p(\textbf{x}|\boldthetahat)$ would collapse to a point mass centered on the desired property value. Then we could sample an input, \mbox{$\xbfhat \sim p(\textbf{x}|\boldthetahat)$}, to obtain our ``designed" input $\xbfhat$. If the desired property referred to a non-infinitesimal set of property values, or if the oracle was not noise-free, then the generator would in general yield a non-degenerate distribution over inputs that we could sample from. This non-degeneracy is intuitively satisfying: it elucidates that the precision to which we can accurately perform input design depends directly on the precision of the oracle.

There are two main difficulties in optimizing \eqref{eq:marginal}. The first is that the parameter being optimized appears in the expectation distribution, which causes difficulties in computing gradients~\cite{KingmaW13, RezendeMW14}. One could use the log-derivative trick to overcome this issue; however, this is known to generally yield high-variance MC estimates. Alternatively, one might consider using the reparameterization trick~\cite{KingmaW13, RezendeMW14}; however, it can be difficult to generally construct reparameterizations, especially for complex distributions such as those encoded by a VAE. Moreover, reparameterization becomes harder still for distributions over discrete state spaces~\cite{Jang2017}. The second difficulty arises from the fact that $S$ will in general be a rare condition, and therefore $P(S|\boldtheta)$ will generally be vanishingly small for most $\boldtheta$, thereby yielding high-variance MC estimates, and exacerbating the first difficulty. We therefore take a different approach.

Next we outline a series of steps that will allow us to tackle the optimization problem in \eqref{eq:marginal} using an iterative algorithm. Specifically, we will find a series of $\boldthetat$, where at each iteration, $t$, the generative model $p(\xbf|\boldthetat)$ gets closer and closer to the desired distribution. Note that conditioning this generative model on $S$ and using Bayes' Theorem we obtain
\begin{align}
    p(\textbf{x}|S, \boldthetat)
    =  \dfrac{P(S|\textbf{x}) p(\textbf{x}|\boldtheta^{(t)})}{P(S|\boldthetat)},
    \label{eq:cond1}
\end{align}
which we can use to derive a lower bound on our objective \eqref{eq:marginal}. As we shall see, we can satiate this bound, so the bound is simply a technique to transform our problem into one we can solve. In particular, we can derive a lower bound as follows:
\begin{align}
    &\log \mathbb{E}_{p(\textbf{x}|\boldtheta)}\left[ 
    P(S |\textbf{x})\right] \label{eq:objective_for_gap}\\
    =&\log \mathbb{E}_{p(\textbf{x}|S, \boldtheta^{(t)})}\left[\dfrac{p(\textbf{x}|\boldtheta)}{p(\textbf{x}|S, \boldtheta^{(t)})} 
    P(S |\textbf{x})\right] \label{eq:lb1} \\
    =& \log \mathbb{E}_{p(\textbf{x}|S, \boldtheta^{(t)})}\left[\dfrac{p(\textbf{x}|\boldtheta)P(S|\boldtheta^{(t)})}{p(\textbf{x}|\boldtheta^{(t)})}\right] \label{eq:lb2} \\
    \geq& \mathbb{E}_{p(\textbf{x}|S, 
     \boldtheta^{(t)})}\left[\log 
    \dfrac{p(\textbf{x}|\boldtheta)}
    {p(\textbf{x}|\boldtheta^{(t)})}\right] + \log P(S|\boldtheta^{(t)})\label{eq:lowerbound}
\end{align}
where \eqref{eq:lb1} introduces $p(\textbf{x}|S, 
\boldtheta^{(t)})$ as an Importance Sampling proposal distribution \cite{MacKay:2002:ITI:971143}, \eqref{eq:lb2} is due to the application of the relationship in \eqref{eq:cond1} and the final lower bound
is due to an application of Jensen's Inequality.
 Also, as shown in the Supplementary Information, the gap between \eqref{eq:objective_for_gap} and \eqref{eq:lowerbound} is given precisely by $D_{KL}\left(p(\xbf|S, \boldtheta^{(t)})||p(\xbf |S,\boldtheta)\right)$, which can be driven to zero because the distributions being compared are in the same parametric family.

Maximizing \eqref{eq:lowerbound} with respect to $\boldtheta$ in order to obtain $\boldthetatplus$ results in the objective:
\begin{align}
    \begin{split} 
    &\argmax_{\boldtheta} \mathbb{E}_{p(\textbf{x}|S, 
     \boldtheta^{(t)})}\left[\log 
    \dfrac{p(\textbf{x}|\boldtheta)}
    {p(\textbf{x}|\boldtheta^{(t)})}\right]+ \log P(S|\boldtheta^{(t)})
    \end{split} 
    \\
    &= \argmax_{\boldtheta} \mathbb{E}_{p(\textbf{x}|S, 
     \boldtheta^{(t)})}\left[\log 
    p(\textbf{x}|\boldtheta) - \log p(\textbf{x}|\boldtheta^{(t)}) \right] \\
    &= \argmax_{\boldtheta} \dfrac{1}{P(S|\boldtheta^{(t)})} \mathbb{E}_{p(\textbf{x}|\boldtheta^{(t)})}\left[ 
    P(S|\textbf{x})
    \log p(\textbf{x}|\boldtheta)\right]\\
    &=\argmax_{\boldtheta} 
    \mathbb{E}_{p(\textbf{x}|\boldtheta^{(t)})}\left[ 
    P(S|\textbf{x})
    \log p(\textbf{x}|\boldtheta)\right].
    \label{eq:obj}
\end{align}
We could in principle begin to optimize  this objective by drawing samples 
from $p(\xbf|\boldtheta^{(t)})$ to calculate a Monte Carlo (MC) estimate of the 
expectation in \eqref{eq:obj}. However, in the case that the $S$ is a rare 
condition, such as when it refers to a maximization or specification condition, then with high probability $P(S|\xbf)$ will be vanishingly small for any $\xbf$ drawn from an initial model.\footnote{If not, then the problem was an easy one that we would have already nearly solved.} Consequently, the MC estimate of the expectation in \eqref{eq:obj}, and its derivative, will exhibit extremely high variance and thus be of little utility in helping us to achieve our goal.

To overcome this difficulty, we will construct a series of 
property value sets, $S^{(t)}$, for 
$t=1,2,\ldots$, such that (a) each $P(S^{(t)}|\boldtheta^{(t)})$ is non-vanishing---that is, sampling from $p(\textbf{x}|\boldthetat)$ will be likely to produce $\xbf$'s such that $P(S^{(t)}|\textbf{x})$ is non-vanishing---and, (b) $S^{(t)}$ approaches $S$ as $t$ grows large. At each iteration, we update the parameters $\boldthetat$ by optimizing an MC estimate of a relaxed version of \eqref{eq:obj} in which $S$ is replaced by $S^{(t)}$. Specifically, we draw $M$ samples, $\xbf_i^{(t)}$, from $p(\textbf{x}|\theta^{(t)})$ and perform the optimization
\begin{equation}
    \boldthetatplus = \argmax_{\boldtheta} \sum_{i=1}^M P(S^{(t)}|\textbf{x}_i^{(t)}) \log p(\xbf_i^{(t)}|\boldtheta) \label{eq:mcobj},
\end{equation}
which amounts to a weighted maximum likelihood (ML) objective which can be optimized with any number of standard techniques for training generative models. The weight of each data point, $\xbf_i^{(t)}$, is given by $P(S^{(t)}|\textbf{x}_i^{(t)})$, which, notably, depends on the uncertainty of the oracle. 
\cut{Note that if we have set $S^{(t)}$ to satisfy the aforementioned criteria, then by construction, the $P(S^{(t)}|\xbf_i^{(t)})$ will tend to have non-vanishing values.}
Note that this is equivalent to minimizing the sample-based KL divergence between the previous generative model, conditioned on $Y\in S^{(t)}$, and the one we are updating, $D_{KL}\left(p(\xbf|S^{(t)}, \boldtheta^{(t)})||p(\xbf|\boldtheta)\right)$.

The only remaining issue is how to set $S^{(t)}$ such that conditions a) and b) above are satisfied.
\cut{\eqref{eq:mcobj} is a low variance MC estimate, and such that $S^{(t)}$ tends towards $S$ as more iterations are performed.} Next we outline the procedure for satisfying these in the maximization context,
and leave the specification problem to the Supplementary Information.
Suppose we wanted to find the $\xbf$ that maximized the expected value of some property, with that maximum value, unknown. Then one would set $S^{(t)}$ to be the relaxed condition that $p(y \geq \gamma^{(t)}| \xbf)$ where $\gamma^{(t)}$ is the $Q^\text{th}$ percentile of property values predicted for those samples from $p(\xbf|\boldtheta^{(t)})$. For $Q$ small enough, $p(S^{(t)}|\xbf)=p(y \geq \gamma^{(t)}| \xbf)$ will be non-vanishing by definition. Thus, by construction, we can now reasonably perform maximization of the objective in \eqref{eq:obj} instantiated with $S^{(t)}$ because the rare event is no longer rare.

We can begin this procedure with some initial model $p(\xbf|\boldtheta^{(0)})$ and iterate until convergence, enforcing that $\gamma^{(t)}$ implied by $Q$ is non-decreasing~\footnote{To guarantee that $\gamma^{(t)}$ is increasing, one would have to sample from the generative model at this time step, until this occurred.}---this can be trivially achieved by  re-using the previous $\gamma^{(t)}$ if needed. Upon convergence, one can draw $M$ samples from $p(\textbf{x}|\boldtheta^{(t)})$, keeping the one with the largest expected value of the property as the designed input. Note that in our empirical evaluations, we instead maintain all $M$ samples in order to better characterize the procedure's performance. 

Convergence can be defined in several ways, including (1) the change in $\boldtheta^{(t)}$ is below some threshold, or (2) the maximum value achieved has not changed after $T$ iterations. However, in our experiments, we instead use a fixed ``sequence budget", described later, enabling fair comparisons between approaches.

Note that if the oracle is noiseless, then the weights collapse to be binary, in which case one should choose $Q$ that is relatively low, such that more samples are given non-zero weight, and correspondingly, the effective sample size---the sum of the weights---is high enough. As the noise of the oracle increases, the weights become increasingly less binary and higher $Q$ can be efficient. Notably, as shown in the Supplementary Information, performance is not particularly sensitive to the choice of $Q$; in fact, $Q \geq 0.4$ worked well across both the noise-free and noisy oracle experiments. Because $Q$ is only defined for $Q \in [0,1]$), so this encompasses nearly the entire range.

The choice of $M$ is related to $Q$ in the sense that if $Q$ is chosen poorly, one may require a larger $M$. However, we have found our method is not terribly sensitive to this setting, and generally use $M$ between 500 and 1,000. Additionally, in our experiments, this parameter is controlled by way of a ``sequence budget" constraint applied to all methods.

There is a trivial condition under which this procedure is guaranteed to find the global maximum, which depends on properties of the generative model. First, the generative model must be high enough capacity. Second, the generative model must assign non-zero probability to every part of the input space. Given these conditions, if one runs the procedure infinitely long, then the global optimum will be found. While this conditions is of no practical utility in achieving the global optimum, it does suggest that the longer we run our approach, the better we are likely to do. This is in direct contrast to the approaches in ~\cite{Killoran2017, Gomez-Bombarelli2018} which have no such guarantees. It is not clear if ~\cite{Gupta2018} may have such guarantees, or if so, possibly only when using a probabilistic generative model such as a VAE, rather than a GAN.

\cut{
We mentioned that at each iteration we are performing maximum likelihood estimation (MLE) of the generative model parameter, $\boldtheta$, using data from the previous iteration's generative model.} 

\begin{algorithm}[ht]
\caption{{\bf Maximization of a single, continuous property.} 
\mbox{$h_\text{oracle}(\xbf_i)$} is a function returning the expected value of the property oracle (\eg the the output of a standard neural network).
\mbox{$\text{CDF}_\text{oracle}(\xbf, \gamma$)} is a function to compute the CDF of the oracle predictive model for threshold $\gamma$ (\eg by setting the neural network variance parameter to the mean-squared error). 
$\text{GenTrain}(\{(\xbf_i, w_i)\}$ is a procedure to take weighted training data $\{(\xbf_i, w_i)\}$, and return a trained generative model. $Q$ is a parameter that determines the next iteration's relaxation threshold; $M$ is the number of samples to generate at each iteration. [$\initset$] is an optional, initial set of samples with which to initialize the generative model. See main text for convergence criteria. Any line with an $i$ subscript implicitly denotes $\forall i \in [1 \ldots M]$.
}
\label{ourmethod}
\begin{algorithmic}
\Procedure{$\ourmethod$}{$h_\text{oracle}(\xbf)$, CDF$_\text{oracle}(\xbf, \gamma)$, $\text{GenTrain}(\{\xbf_i\, w_i\})$, [$Q=0.9$], [$M=1000$], [$\initset = \emptyset$]} 
\If{$\{\xbf_\text{init}\}$ is empty} 
\State $set \gets \initset$
\Else \State $set_i \gets \text{randomly initialize } \xbf_i$
\EndIf
\State $weights_i \gets \text{1.0, }$
\State $\gamma^0= \text{median}(\{h_\text{oracle}(\xbf_i)\})$ 
\State $t \gets 1$
\While {\text{not converged}}
    \State $G \gets \text{GenTrain(} set, weights \text{)}$
    \State $set_i \gets \xbf_i \sim G$
    \State $scores_i \gets h_\text{oracle}(\xbf_i)$
    \State \mbox{$c\_scores_i \gets 1-CDF_\text{oracle}(\xbf_i, \gamma^{(t-1)})$}
    \State $q \gets \text{index of } Q^\text{th} \text{ percentile of } c\_scores$
    \State $\gamma^{(t)} \gets scores_\text{q}$
    \State \mbox{$weights_i \gets 1-CDF_\text{oracle}(\xbf_i, \gamma^{(t)})$}
    \State $t \gets t+1$
\EndWhile
\Return $set, weights$
\cut{
\BState \emph{top}:
\If {$i > \textit{stringlen}$} \Return false
\EndIf
\State $j \gets \textit{patlen}$
\BState \emph{loop}:
\If {$\textit{string}(i) = \textit{path}(j)$}
\State $j \gets j-1$.
\State $i \gets i-1$.
\State \textbf{goto} \emph{loop}.
\State \textbf{close};
\EndIf
\State $i \gets i+\max(\textit{delta}_1(\textit{string}(i)),\textit{delta}_2(j))$.
\State \textbf{goto} \emph{top}.
}
\EndProcedure
\end{algorithmic}
\end{algorithm}

\paragraph{Connections to other methods}

There is an interesting connection between \ourmethod\space and an approach called the Cross Entropy Method
(CEM)~\cite{Rubinstein1999,Rubinstein1997}. CEM is a method originally developed to estimate 
the probability of rare events, and then later generalized to tackle optimization. CEM's 
derivation stems from a goal of achieving the lowest variance estimate of a rare event probability, 
which yields an objective similar to that in \eqref{eq:obj}. CEM also bears other similarities 
to our approach, although, to the best of our knowledge, it has not been used to solve the 
design problem for stochastic oracles, nor has it been been combined with powerful modern-day 
generative models. 
In fact, in some sense, one can view \ourmethod\space as the natural extension of
CEM to our problem statement, even though we arrived at it from a completely different 
angle---one befitting the problem statement. CEM in turn has a strong connection to 
reward-weighted regression used in Reinforcement Learning (\eg~\cite{Peters2007, Hachiya2011}).

Another interesting connection lies between \ourmethod\space and Expectation-Maximization (EM). At
each iteration, when we optimize \eqref{eq:obj}, we are solving a weighted MLE problem, much
like in EM. In our approach, the samples are weighted by $P(S|\xbf)$, whereas in EM they are
weighted by the posterior distribution over the latent variable. Thus one can interpret 
$P(S|\xbf)$ as a posterior distribution over whether our desideratum is satisfied or not. 
Analogously to EM where the posterior moves 
closer and closer to the true posterior, in \ourmethod, the approximate ``posterior" 
distribution slowly moves toward the true posterior as $S^{(t)} \rightarrow S$, albeit using something more akin to annealing. 

Our method also bears resemblance to the Covariance Matrix Adaptation Evolution Strategy (CMA-ES) \cite{CMA-ES}. However, CMA-ES can only use a multivariate Gaussian as its underlying generative model, which limits its applicability to many problem areas.

Dayan and Hinton \cite{Dayan1997UsingEF} used a similar lower bound to ours to show that EM can be used to increase the expected reward in a particular Reinforcement Learning algorithm.

\cut{
 ABC
4.	Annealed Importance Sampling?
5.	Combinatorial optim. And other approaches don’t make use of uncertainty in the oracle, which we do.
}

\section{EXPERIMENTS}

\begin{figure*}[t]
    \centering
    \begin{subfigure}[t]{0.45\textwidth}
        \centering
        \includegraphics[width=\textwidth]{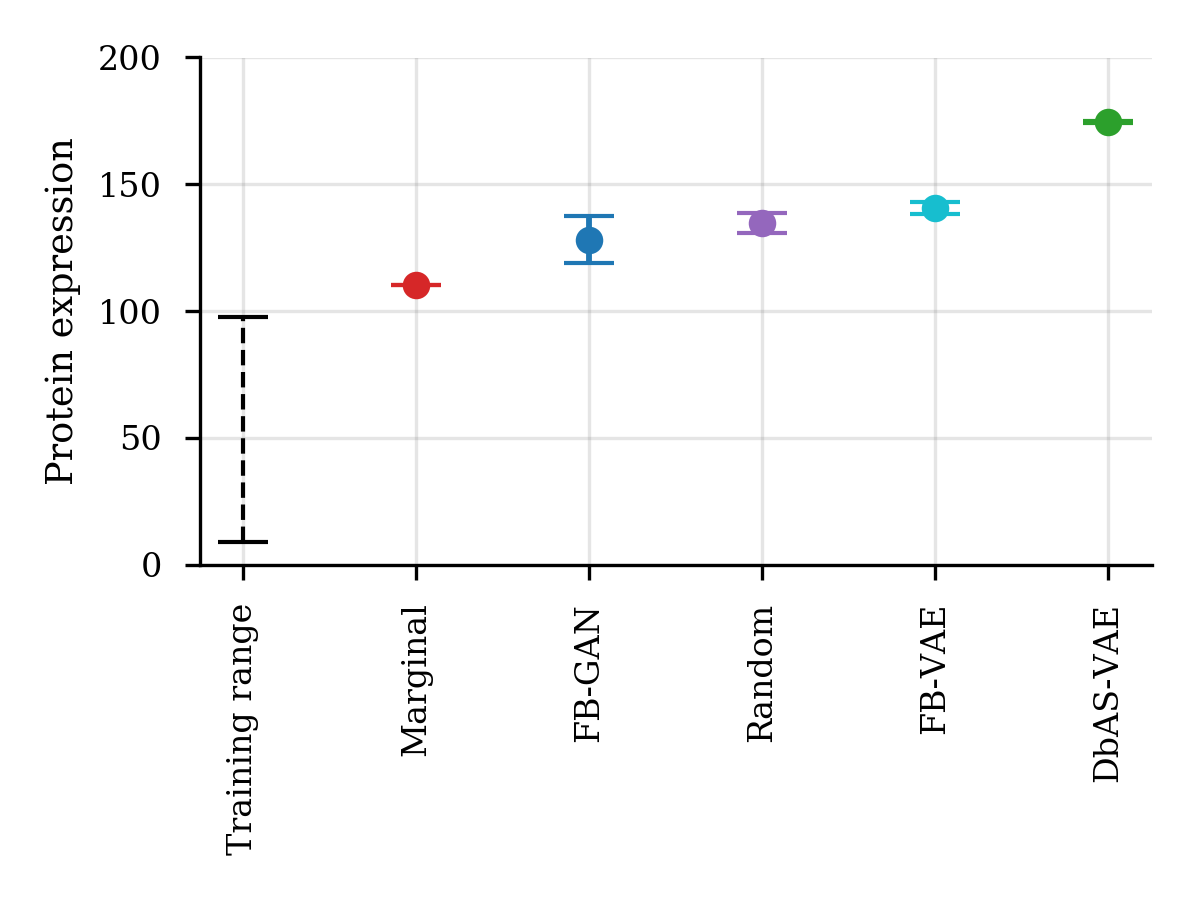}
        \caption{}
    \end{subfigure}
    \begin{subfigure}[t]{0.45\textwidth}
        \centering
        \includegraphics[width=\textwidth]{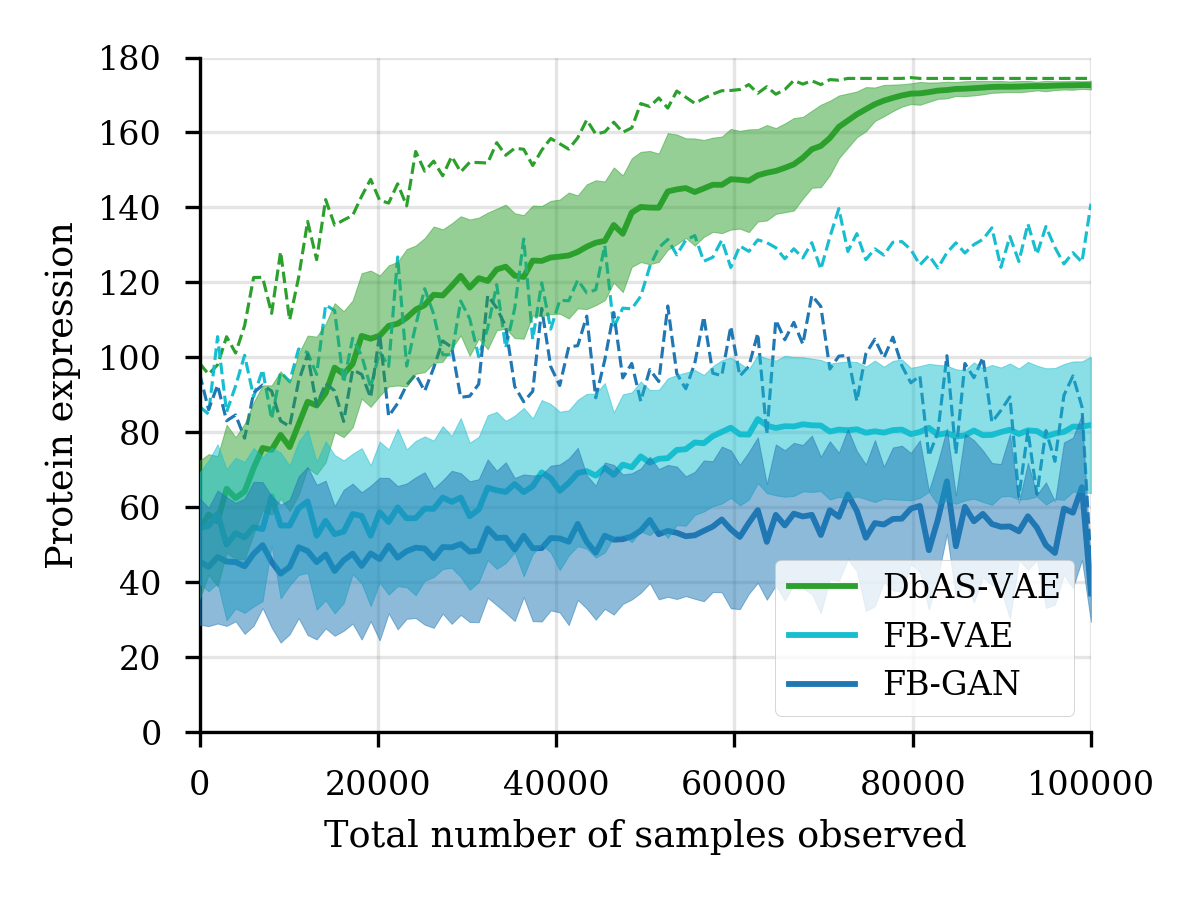}
        \caption{}
    \end{subfigure}
    \caption{Maximization of a noisy, protein expression oracle. (a) Mean $\pm$ one standard deviation of the protein expression over five runs for each method. The dashed line represents the range of expression values in the training set. Killoran is not shown because it could not accommodate design constraints to produce the correct protein, (b) Representative trajectories for iterative methods. Plot style follows that of Figure \ref{fig:figure1}b.}
    \label{fig:figure2}
\end{figure*}

\paragraph{Overview}
We conduct three sets of proof-of-concept experiments that demonstrate the capabilities
of \ourmethod. In all cases, we focus on designing sequences with respect to a single property.
In the first set we generate random, noiseless oracle functions for a single property we want to maximize. We chose this random function setting to ensure that our approach could work in a general sense not specific to any domain. Moreover, we restrict the experiments to sequences that are short enough that we can enumerate all of them, and therefore know the ground truth maximum (sequence length $L=13$ corresponding to \textapprox$10^8$ possible sequences). 

In the second set of experiments, our goal is to examine the feasibility of using our method in a particular domain of interest---maximizing the protein yield by designing DNA sequences. To do so, we learn a noisy oracle model on a large-scale  set of protein expression data~\cite{Cambray2018} (Supplementary Figure 1), and then use this model to generate simulated data on which to evaluate the different methods. Here the design space is too large to search through and so we compare the property values achieved by each method, without knowledge of the true maximum (sequence length $L=96$ corresponding to \textapprox$10^{16}$ possible sequences). In the third set of experiments, we again use the protein yield simulation set-up, but this time for the specification problem.
In all of these experiments, $\xbf$ represents a DNA sequences of length $L$ (containing $L\times 4$ entries from the one-hot encoding of nucleotides, $\{A,C,T,G\}$), and $y \in \mathbb{R}$ represents a scalar property of the sequences that we want to maximize or specify.

\paragraph{Method comparison}
In addition to \ourmethod-VAE---an implementation of \ourmethod~that uses a VAE as the generative model---we also compare to: (1) {\it Random} design, where a given number of sequences are independently and identically sampled from a discrete distribution assigning equal mass to each discrete possibility, (2) {\it Marginal} design, wherein we design an optimal sequence by setting each position in the sequence to the element with the highest average oracle value in the training set, 
\cut{(3) a depth-1 greedy search through the design space (\ie~at each iteration, use the oracle to score every sequence which is within edit distance of one and choose the best),} 
(3) the method of Killoran \etal~\cite{Killoran2017} (with a GAN prior), (4) {\it FB-GAN}, as described in \cite{Gupta2018}, and (5) {\it FB-VAE}, a version of FB-GAN that instead uses the same VAE architecture as in \ourmethod, so that we can isolate the differences in performance owing to the design procedure rather than the choice of generative model. Further details about these methods are in the Supplementary Information. 

To fairly compare these methods we keep the total number of samples considered during each run, $N$, constant. We call this the sequence budget. This sequence budget corresponds to limiting the number of total gradient step updates performed in the Killoran method; limiting the total number samples drawn from the generative model in FB-GAN, FB-VAE, and \ourmethod; and limiting the total number of sequences that the Random method is allowed to consider\cut{; and limiting the number of search steps the greedy search can take}. The Marginal method operates outside of these constraints because it only ever considers sequences in the training set, and requires no budget limitations. Additionally, every method is provided with the same initial training sets for each sequence, except for Random, which requires no training set. 

Throughout, the term {\it training data} refers to sequences that each method is seeded with, and which are not paired with labels/properties. For \ourmethod, FB-GAN, FB-VAE, and Killoran, these are used to train the unsupervised generative model at the first iteration. For Marginal, it is used to compute the marginal statistics.

\subsection{Maximization of random, noise-free oracles} \label{random_nn_section}

We first test the ability of the methods to maximize the scalar
output of dense neural network functions with randomly assigned weights. 
\cut{The search space in these problems is all possible DNA sequences of a specified length $L$. More specifically, for a given sequence length $L$, we build a neural network oracle with random weights that takes sequence representations as input and outputs a scalar value, and run each of the above methods with the aim of finding the length-$L$ sequence that maximizes the output of the network.}
We run this test for sequences lengths, $L$ between 6 and 13. These sequences are short enough that we can brute-force construct all $4^L$ possible sequences and evaluate their function values in the oracle. This allows us to determine the ability of each method to find the \textit{global} maximum in the search space, in the small design space setting. 

We assume that the oracle is an unbiased, noiseless predictor of the
property. Therefore, within the \ourmethod\space framework, we use \mbox{$p(y|\textbf{x}) = \delta(f(\textbf{x})-y)$}
and correspondingly $P(Y\geq y|\xbf) = \mathbbm{1}_{\{f(X)\geq y\}}(\xbf)$ where $f(\textbf{x})$ 
is the random neural network oracle, and $\delta(f(\textbf{x})-y)$ denotes the Dirac delta function. 
\cut{and $\{f(X)\geq y\}$ is the set of all function evaluations greater than $y$.}
We set $Q=0.95$. However, as mentioned in Methods and shown in the Supplementary Information, \ourmethod~is not very sensitive to this setting. 

For each sequence length, we constructed 10 random training sets each of 1,000 sequences, under the constraint that none of the training sequences corresponded to oracle values above the 40th percentile over all $4^L$ sequences, thereby ensuring that there were no near-optimal sequences in the training set. We then ran each method once on each training data set with sequence budget $N=10,000$.\footnote{In \ourmethod, FB-GAN and FB-VAE the sequence budget was split over 20 iterations \ie the methods sample $M=500$ sequences at each iteration.}

The maximum oracle values found by each method were then compared by calculating a score representing how much of the gap between the highest property value in the training data and the global maximum was achieved:\footnote{Had we just compared the scores directly, the fluctuation across random functions would have obscured the results.}
\begin{equation}
    \textit{fraction of possible gain} = \dfrac{ y_{\text{opt}} - y^*_{\text{train}}}{y^*_{\text{global}} - y^*_{\text{train}}},
\end{equation}
where $y_{\text{opt}}$, $y^*_{\text{train}}$, and $y^*_{\text{global}}$ are the the maximum value found during design, the maximum value in the training set, and the global maximum, respectively. This score achieves its maximum of 1 when a method finds the global optimum.

In this setting, \ourmethod\space nearly always finds the global optimum sequence while FB-GAN, FB-VAE and Killoran methods struggle to perform better than the Random search (Figure \ref{fig:figure1}a). This suggests that the prior used by these methods overly influences and constrains the search. That the Marginal method performs so poorly demonstrates that the oracle contains complex signal. Furthermore, 
Killoran quickly converges to a local maximum, while FB-GAN and FB-VAE move very slowly towards generating sequences with higher property values, and \ourmethod~quickly converges to the global maximum (Figure \ref{fig:figure1}b). 

\cut{
The greedy search performs well, though less consistently than DbAS, in this setting. We posit this is due to a possible lack of local optimum in the random neural network. As we will see in the next experiment, the greedy search performs substantially worse than all other methods in a larger, more realistic design problem.
}

\subsection{Maximization of a noisy, protein expression oracle}\label{expmax}

We next tested the methods on domain-specific, noisy oracles. In particular, on oracles created by fitting a predictive model on protein expression data. The aim in this domain is to find the DNA sequence that yields the most of a specified protein (the protein expression).\footnote{Due to the redundancy in the genetic code, many DNA sequences can code for the same protein, but the expression level of the protein depends on the particular DNA sequence that
coded for it.} The oracle model was given by \mbox{$p(y|\xbf)=\mathcal{N}(y|f(\xbf), \sigma^2)$}, where $f$ is a neural network, trained on sequence-expression pairs from~\cite{Cambray2018}; $\sigma^2$ was set to its maximum likelihood value. The maximum protein expression value in the data set was 100.

An added difficulty in this problem compared to the previous experiment is that the design problem is now constrained to produce a DNA sequence corresponding to a specific protein sequence. To incorporate this constraint into \ourmethod, we construct an additional, deterministic oracle, $p(T=\boldsymbol{\alpha}|\textbf{x})=\mathbbm{1}_{\{t(X)=\boldsymbol{\alpha}\}}(\xbf)$, where $T$ is a random variable representing translated sequences, $\boldsymbol{\alpha}$ is the target protein, $t$ is a function that translates DNA sequences, and $\{t(X)=\boldsymbol{\alpha}\}$ is the set of sequences that are translated into $\boldsymbol{\alpha}$. 
\cut{For every sample, we multiply the output of this oracle by that of the expression prediction oracle to arrive at a final weight for the sample.} This secondary oracle is used for \ourmethod, FB-GAN and FB-VAE; for the latter two, we naturally generalized the method to accomodate this. The Killoran method cannot incorporate this constraint without substantial changes to the method and we therefore do not test it in this experiment. We set $Q=0.99$, corresponding to the fact that with a noisy oracle, one can effectively use any large $Q$. However, as mentioned earlier and shown in the Supplementary Information, our method is not particularly sensitive to the choice of this parameter.

We chose the protein with the most experimental observations to test for design. The protein had 418 DNA sequence-expression pairs experimentally measured; these were used to initialize the generative model. We performed five runs with each method, with a sequence budget of $N=100,000$ samples.\footnote{This is larger than in the first experiment corresponding to the much larger search space. The budget is split into $M=1000$ samples at each of 100 iterations in \ourmethod, FB-GAN, and FB-VAE} Again, \ourmethod~performs substantially better than the competing methods, and its sample trajectory converges to a predicted optimum value (Figure \ref{fig:figure2}).

\subsection{Specification of noisy, protein expression oracle}\label{expspec}

In our final experiment, we tested the capability of \ourmethod-VAE to perform specification in the same context as in Section \ref{expmax}. 
That is, we used the same protein expression oracle, and set the target set, $S$, to an infinitesimal range around a specific expression value, $c$, that is not the optimum, for the same protein as in the previous section. We chose five evenly-spaced values of $c$, starting at 10 and going up to 130. We used the same settings for $Q$, the sequence budget, and number of iterations as in the previous section.

\begin{figure}[h]
        \includegraphics[width=1\linewidth]{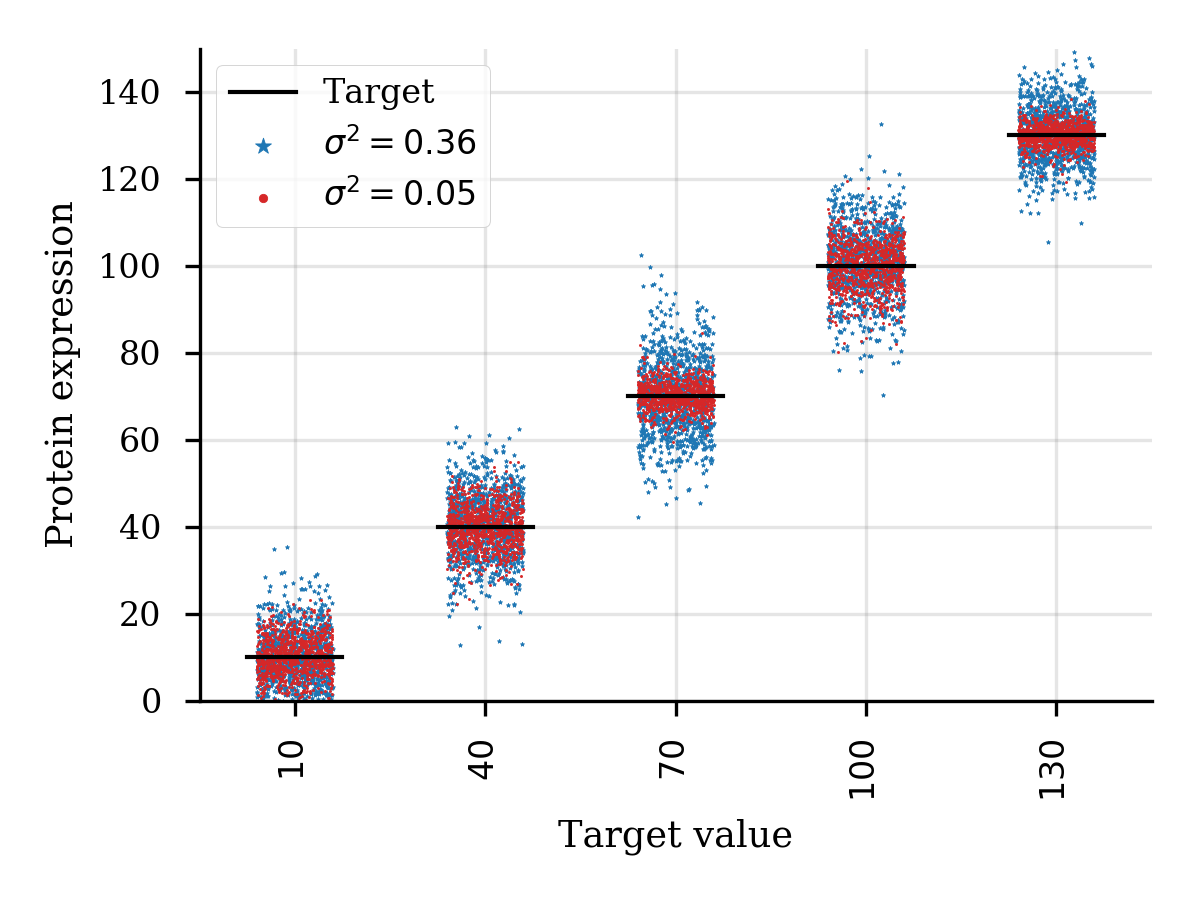}
        \caption{Expression specification experiment results. Each block of points 
        corresponds to experiments with a specified expression value, denoted by the value on the vertical axis of the solid black lines. 
        Points shown within a block are 1000 samples taken from the VAE in the final iteration of a \ourmethod\space run.
        The vertical axis shows the oracle predicted expression values. 
        The horizontal axis positions for samples corresponding to a given target value are uniformly, randomly placed for ease of visualization. 
        Experiments were run with two settings of the oracle variance, one with
        the maximum likelihood variance (blue stars), and one with artificially lowered variance (red circles). }
        \label{fig:figure3}
        \vspace{-0.15in}
\end{figure}

\ourmethod~is effective at concentrating the generative model around the target value, albeit with significant 
variance of sampled values around the target (Figure \ref{fig:figure3}). We hypothesize that this is partially owing to the variance of the oracle predictions, and confirmed this by artificially reducing the oracle variance from its maximum likelihood value of 0.36 to 0.05 and repeating the experiment. As expected, this produces tighter distributions around the specified value. We hypothesize that the remaining variance in sampled values is likely due to the well-known ``blurriness" problem in VAEs \cite{Engel2018}, which prevents the generative model from collapsing around sequences whose expected predicted expression are closest to the specified value.

 \section{DISCUSSION}
 
 We have presented a new state-of-the-art method, \ourmethod, for designing inputs to satisfy certain properties, which can be evaluated, noisily or not, with an oracle model. When the oracle is noisy, with known noise properties, our method can leverage this noise. In our experiments, we assumed homoscedastic oracle noise. However, it will be important to develop and use oracle models that have heteroscedastic noise, such as GP regression~\cite{rasmussen2004}, or developments to achieve similar capabilities in neural networks~\cite{Gal2016, calibration2017}. Additionally, it will be important to understand more deeply how issues of interpolation versus extrapolation may affect the ability to design well.
 
 Although our experiments focused on designing discrete sequences with real-valued properties, \ourmethod readily handles real-valued design spaces, and discrete properties (\ie classifier oracles). We also focused on VAE as the plug-and-play choice for the generative model, but any other generative model that can be trained using weighted MLE can be used.

\cut{ 
One issue that deserves further attention is restricting the design space to feasible regions which cannot be defined by the oracles. In our experiments the entire design space was feasible, such as would occur in optimizing over DNA space to optimize protein yield (one can trivially restrict the DNA space to only the desired protein). However, if one wanted to design say images which adhered to properties of natural images, such as smoothness with occasional edges, then it could very well be sufficient to seed the \ourmethod\space generative model with a collection of images which satisfy these properties (but importantly, need not capture the maximization/specification property); as \ourmethod\space iterates, it does so only smoothly moving the generative distribution toward the desired properties. \ourmethod\space could end up keeping the generative distribution within the originally seeded region, although there would be no guarantee. To be certain of this, one could build a hierarchical generative model in which one first learns the manifold of the feasible region within the entire design space given a set of feasible examples, and then performs \ourmethod\space within this fixed manifold. Alternatively, one could impose a feasibility constraint in a manner similar to the realism constraint implemented in ~\cite{Engel2018}.

Another direction to explore is replacing the KL divergence that \ourmethod\space uses when updating the generative model. For example, one may want to explore a symmetrized KL divergence or others divergences altogether. The choice of divergence will change the way in which \ourmethod\space moves through the design space. However, it is likely that the KL divergence will be by far the easiest to implement because it yields a weighted MLE problem, and hence will likely remain the divergence of practical choice.
}

Because the generative model is updated at every step, our approach to design can in principle be applied to the ``zero-shot" problem, wherein no generative model training data at all are provided. However, we have not yet investigated this setting because most competing methods require training data.

A direction we are actively pursuing is to leverage \ourmethod\space to do rare-event, or in the extreme case, zero-shot conditional generative modelling. Such a goal would be similar to that in~\cite{Engel2018}, however, our approach would provide a coherent framework within which to do it.

Finally, we are currently pursuing use of \ourmethod\space for an end-to-end protein design problem that includes wet-lab experimental validation.


\subsubsection*{Acknowledgements}
The authors thank Sergey Levine, Benjamin Recht, Prasad Raghavendra, Nicolo Fusi, Lester Mackey and Tina Liu for discussion, Kevin Murphy for pointers to the CMA-ES literature, and Quaid Morris for help in choosing the title and acronym.

This research used resources of the National Energy Research
Scientific Computing Center, a DOE Office of Science User Facility
supported by the Office of Science of the U.S. Department of Energy
under Contract No. DE-AC02-05CH11231

\bibliographystyle{abbrv}
\bibliography{paper}
\clearpage

\setcounter{figure}{0}
\setcounter{section}{0}
\setcounter{equation}{0}
\renewcommand{\thefigure}{S\arabic{figure}}
\renewcommand{\thesection}{S\arabic{section}}
\renewcommand{\theequation}{S\arabic{equation}}
\aistatstitle{Supplementary Information: Design by adaptive sampling}
\thispagestyle{empty}

\section{Generalization to specification}

The \ourmethod~ procedure can be easily extended to perform specification of a property value rather than maximization. In this case the target set is an infinitesimally small range around the target, \ie $S = [y_0 - \epsilon, y_0 + \epsilon]$ for a target value $y_0$ and a small $\epsilon > 0$.  The DbAS procedure remains mostly identical to that of maximization case, except in this case the intermediate sets $S^{(t)} = [y_0 - \gamma^{(t)}, y_0 + \gamma^{(t)}]$ are centered on the specified value and have an update-able width, $\gamma^{(t)}$. The $\gamma^{(t)}$ values are then updated analogously to the thresholds in the maximization case, \ie $\gamma^{(t)}$ is set to the $Q^\text{th}$ percentile of $|y_i - y_0|$ values, for $i=1,...,M$, where $y_i$ are the expected property values according to the sample $\textbf{x}_i \sim p(\textbf{x}|\boldthetat)$.
\section{Generalization to multiple properties}

Additionally, DbAS can be extended to handle multiple properties $Y_1,...,Y_K$ with corresponding desired sets $S_1,...,S_K$. We only require that these properties are conditionally independent given a realization of $X$. In this case, 
\begin{equation}
    P(S_1,...,S_K|\textbf{x}) = \prod_{i=1}^K P(S_i|\textbf{x})
\end{equation}
where now each $Y_i$ has an independent oracle. This distribution, and the corresponding marginal distribution $P(S_1,...,S_K|\boldtheta) = \int d\textbf{x}\, p(\textbf{x}|\boldtheta) \prod_{i=1}^K P(S_i|\textbf{x})$ can then be used in place of $P(S|\xbf)$ and $P(S|\boldtheta)$ in Equations (1)-(13) in the main text to recover the DbAS procedure for mutiple properties.

\section{Alternative derivation of lower bound}

Here we provide an alternative derivation of the lower bound in Equation (8) of the main text. 
We begin by considering the quantity $D_{KL}\left(p(\xbf|S, \boldtheta^{(t)})||p(\textbf{x}|S,\boldtheta)\right)$ (where $D_{KL}$ is the KL-divergence between two distributions), 
which will ultimately be the difference between the right- and left-hand sides of the bound (similar in spirit to the difference that emerges from the lower bound used in variational inference ~\cite{Jordan1999}):
\begin{align}
    &D_{KL}\left(p(\xbf|S, \boldtheta^{(t)})||p(\xbf |S,\boldtheta)\right) \label{eq:lb_dkl}\\
    =& \mathbb{E}_{p(\xbf|S, \boldtheta^{(t)})}\left[\log p(\xbf |S, \boldtheta^{(t)}) - \log p(\xbf |S, \boldtheta)\right] \label{eq:lb_alt_1} \\
    =& \mathbb{E}_{p(\xbf|S, \boldtheta^{(t)})}\left[\log \dfrac{p(S|\xbf)p(\xbf|\boldtheta^{(t)})}{P(S|\boldtheta^{(t)})}-\log \dfrac{P(S|\xbf)p(\xbf|\boldtheta)}{P(S|\boldtheta)}\right] \label{eq:lb_alt_2}\\
    =& -\mathbb{E}_{p(\xbf|S, \boldtheta^{(t)})}\left[\log \dfrac{p(\xbf|\boldtheta)}{p(\xbf|\boldtheta^{(t)})}\right] - \log P(S|\boldtheta^{(t)}) + \log P(S|\boldtheta),
    \label{eq:lb_alt_3}
\end{align}

where \eqref{eq:lb_alt_1} is due to the definition of KL-divergence, \eqref{eq:lb_alt_2} is due to an application Bayes' Theorem to both terms in the expectation, and \eqref{eq:lb_alt_3} is due to the cancellation of the $\log P(S|\xbf)$ terms and a rearrangement of terms. Rearranging the equality between \eqref{eq:lb_dkl} and \eqref{eq:lb_alt_3}, we arrive at:
\begin{align}
    &\log P(S|\boldtheta) - D_{KL}\left(p(\xbf|S, \boldtheta^{(t)})||p(\xbf |S,\boldtheta)\right) \label{eq:lb_eq1} \\
    =& \mathbb{E}_{p(\textbf{x}|S, \boldtheta^{(t)})}\left[\log \dfrac{p(\textbf{x}|\boldtheta)}
    {p(\textbf{x}|\boldtheta^{(t)})}\right] + \log P(S|\boldtheta^{(t)}) \label{eq:lb_eq2}.
\end{align}
\onecolumn
Again, similarly to how the Evidence Lower Bound in variational inference can be developed~\cite{Jordan1999, BleiVarTut},
the equality in \eqref{eq:lb_eq2}, combined with the fact that the KL-divergence is always non-negative,
implies the lower bound in (8). Additionally, this equality makes it clear that by maximizing (8) according to the objective in Equation (12), that we are simultaneously maximizing $\log P(S|\boldtheta)$ and minimizing $D_{KL}\left(p(\xbf|S, \boldtheta^{(t)})||p(\xbf |S,\boldtheta)\right)$. The two densities in this KL divergence term have the same parametric form, and therefore this divergence can be driven to zero.

\section{Extension to models not permitting MLE}

Many models cannot be fit with maximum likelihood estimation, and in this case we cannot solve the DbAS update equation, \eqref{eq:mcobj}, exactly. However, DbAS can still be used in the case when approximate inference procedures can be performed on these models, for example any model that can be fit with variational inference \cite{Jordan1999}. We derive the DbAS update equation in the variational inference case below, but the update equation can be modified in a corresponding way for 
any model that permit other forms of approximate MLE.

In variational inference specifically, the maximum likelihood is lower bounded by an alternative objective:
\begin{align}
    &\max_{\boldtheta} \log p(\xbf|\boldtheta)
    \\= &\max_{\boldtheta} \log \mathbb{E}_{p(\textbf{z})}[p(\xbf|\textbf{z})] \\ 
    \geq &\max_{\boldtheta, \boldsymbol{\phi}} \mathbb{E}_{q(\textbf{z}|\xbf, \boldsymbol{\phi})}\left[\log p(\xbf|\textbf{z}, \boldtheta)\right] - D_{KL}\left[q(\textbf{z}|\xbf, \boldsymbol{\phi})||p(\textbf{z})\right]  \\
    = &\max_{\boldtheta, \boldsymbol{\phi}} \mathcal{L}(\textbf{x}, \boldtheta, \boldsymbol{\phi})\label{eq:elbobound}
\end{align}
where $\textbf{z}$ is a realization of a latent variable with prior $p(\textbf{z})$, and $q(\textbf{z}|\xbf, \boldsymbol{\phi})$ is an approximate posterior with parameters $\boldsymbol{\phi}$.  Equation \eqref{eq:elbobound} implies that we can lower bound the the argument of \eqref{eq:mcobj}:
\begin{equation}\label{eq:gencem_update3}
    \max_{\boldtheta }\sum_{i=1}^M  P(S^{(t)} |\textbf{x}_i^{(t)}) \log p(\textbf{x}_i^{(t)}|\boldtheta) \geq \max_{\boldtheta, \boldsymbol{\phi}} \sum_{i=1}^M P(S^{(t)} |\textbf{x}_i^{(t)}) \mathcal{L}(\textbf{x}_i^{(t)}, \boldtheta, \boldsymbol{\phi})
\end{equation}
which is a tight bound when the approximate posterior in the model is rich enough for the approximate posterior to exactly match the true model posterior. This suggests a new update equation, specific for models trained with variational inference:
\begin{equation}
    \boldtheta^{(t+1)}, \boldsymbol{\phi}^{(t+1)} = \argmax_{\boldtheta, \boldsymbol{\phi}} \sum_{i=1}^M p(S^{(t)} |\textbf{x}_i^{(t)}) \mathcal{L}(\textbf{x}_i^{(t)}; \boldtheta, \boldsymbol{\phi})
\end{equation}
where we now give time dependence to the approximate posterior parameters, $\boldsymbol{\phi}$. In practice, this is the update equation we use for DbAS-VAE.

\section{Keeping Old Samples}

We can extend our method to allow one to maintain samples drawn from the generative models at times before the current iteration.  Beginning with the argument to the program in \eqref{eq:obj}:
\begin{align}
    \mathbb{E}_{p(\textbf{x}|\boldtheta^{(t)})}\left[ 
    P(S^{(t)}|\textbf{x})
    \log p(\textbf{x}|\boldtheta)\right] 
    &= \dfrac{1}{t} \sum_{k=0}^{t} \mathbb{E}_{p(\textbf{x}|\boldtheta^{(t)})}\left[ 
    P(S^{(t)}|\textbf{x})
    \log p(\textbf{x}|\boldtheta)\right] \\
    &= \dfrac{1}{t} \sum_{k=0}^{t} \mathbb{E}_{p(\textbf{x}|\boldtheta^{(t)})}\left[ \dfrac{p(\xbf|\boldtheta^{(k)})}{p(\xbf|\boldtheta^{(k)})}
    P(S^{(t)}|\textbf{x})
    \log p(\textbf{x}|\boldtheta)\right] \\
    &= \dfrac{1}{t} \sum_{k=0}^{t} \mathbb{E}_{p(\textbf{x}|\boldtheta^{(k)})}\left[ \dfrac{p(\xbf|\boldtheta^{(t)})}{p(\xbf|\boldtheta^{(k)})}
    P(S^{(t)}|\textbf{x})
    \log p(\textbf{x}|\boldtheta)\right] \\
    & \approx \dfrac{1}{t} \sum_{k=0}^{t} \sum_{i=0}^{M} \dfrac{p(\xbf^{(k)}_i|\boldtheta^{(t)})}{p(\xbf^{(k)}_i|\boldtheta^{(k)})}
    P(S^{(t)}|\xbf^{(k)}_i)
    \log p(\xbf^{(k)}_i|\boldtheta)\label{eq:old_mcobj}
\end{align}
where $\xbf^{(k)}_i$ are samples drawn from $p(\xbf|\boldtheta^{(k)})$ at the $k^\text{th}$ iteration of our method. Equation \eqref{eq:old_mcobj} can be used as an alternative objective to \eqref{eq:mcobj} if one wishes to maintain old samples.

Of course, it may not be possible to calculate $p(\xbf | \boldtheta^{(k)})$ for some models. In particular, $p(\xbf|\boldtheta^{(k)})$ is intractable for many latent variable models. However, the above derivation can be modified slightly to only require the values of $p(\xbf |\textbf{z}. \theta^{(k)})$ which is very easy for models such as a VAE. Note that this does require caching both $\xbf$
and $\textbf{z}$ samples from previous timesteps.
\begin{align}
    \mathbb{E}_{p(\textbf{x}|\boldtheta^{(t)})}\left[ 
    P(S^{(t)}|\textbf{x})
    \log p(\textbf{x}|\boldtheta)\right] 
    &= \dfrac{1}{t} \sum_{k=0}^{t} \mathbb{E}_{p(\textbf{x}, \textbf{z}|\boldtheta^{(t)})}\left[ 
    P(S^{(t)}|\textbf{x})
    \log p(\textbf{x}|\boldtheta)\right] \\
    &= \dfrac{1}{t} \sum_{k=0}^{t} \mathbb{E}_{p(\textbf{x}, \textbf{z}|\boldtheta^{(t)})}\left[ \dfrac{p(\xbf, \textbf{z}|\boldtheta^{(k)})}{p(\xbf, \textbf{z}|\boldtheta^{(k)})}
    P(S^{(t)}|\textbf{x})
    \log p(\textbf{x}|\boldtheta)\right] \\
    &= \dfrac{1}{t} \sum_{k=0}^{t} \mathbb{E}_{p(\textbf{x}, \textbf{z}|\boldtheta^{(k)})}\left[ \dfrac{p(\xbf|\textbf{z},\boldtheta^{(t)})p(\textbf{z})}{p(\xbf|\textbf{z},\boldtheta^{(k)})p(\textbf{z})}
    P(S^{(t)}|\textbf{x})
    \log p(\textbf{x}|\boldtheta)\right] \\
    & \approx \dfrac{1}{t} \sum_{k=0}^{t} \sum_{i=0}^{M} \dfrac{p(\xbf^{(k)}_i|\textbf{z}_i^{(k)
    },\boldtheta^{(t)})}{p(\xbf^{(k)}_i|\textbf{z}_i^{(k)}, \boldtheta^{(k)})}
    P(S^{(t)}|\xbf^{(k)}_i)
    \log p(\xbf^{(k)}_i|\boldtheta)\label{eq:old_mcobj}
\end{align}

\section{Experimental Details}

Here we provide the necessary details to run the experiments described in the main text. In what follows, when we specify model architectures we use the notation \texttt{LayerType(OutputShape)} to describe layers, and the notation $\texttt{Layer1(Out1)} \rightarrow \texttt{Layers2(Out2)}$ to denote that $\texttt{Out1}$ is given as the input to $\texttt{Layer2}$.

\subsection{Methods details}

\paragraph{VAE architecture} The following VAE architecture was used for all DbAS-VAE and FB-VAE experiments. The VAE encoder architecture is $\texttt{Input(L, 4)} \rightarrow \texttt{Flatten(L*4)} \rightarrow \texttt{Dense(50)} \rightarrow \texttt{Dense(40)}$. The final output is split into two vectors of length 20, which represent the mean and log-variance of the latent variable, respectively. The decoder architecture is $\texttt{Input(20)} \rightarrow \texttt{Dense(50)} \rightarrow \texttt{Dense(L*4)} \rightarrow \texttt{Reshape(L, 4)} \rightarrow \texttt{ColumnSoftmax(L, 4)}$.

\paragraph{FB-GAN and FB-VAE parameter settings} A major implementation choice in 
FB-GAN and FB-VAE is the value of the threshold used to 
decide whether to give 0/1 weights to samples. We found that setting the threshold to the 
$80^{\text{th}}$ percentile of the property values in the initial training set gave the best 
performance, and used that setting for all tests presented here.

\paragraph{FB-VAE implementation} We note that a minor modification to the FB-GAN framework was required to accomodate a VAE generator instead of a GAN. Specifically we must have the method sample from the distribution 
output by the VAE decoder in order to get sequence realizations, rather than taking the 
argmax of the Gumbel-Softmax approximation output by the WGAN.

\paragraph{GAN architectures} Both the Killoran and FB-GAN methods use the WGAN architecture originally proposed in \cite{DBLP:journals/corr/GulrajaniAADC17}, with modifications to allow
for discrete generator outputs. Specifically, the Killoran method applies a softmax layer and FB-GAN a
Gumbel-softmax layer (with $\tau=0.75$), to the final layer of a neural network generator that outputs 
any $L \times 4$ matrix. Both of these produce continuous approximations to discrete random variables, and thus allow one to take gradients with respect to the generator parameters. We employ these architectures for all tests of these methods, with the dimensionality of the latent space set to 100, as suggested by Killoran, \etal. (Note that Gupta \etal\, do not provide a setting for the dimensionality of the latent space). We note that in the expression maximization experiment (Section \ref{expmax}), we found that the WGAN with Gumbel-Softmax used by FB-GAN produced almost no samples that corresponded to the target protein after training with the 418 experimental sequences that correspond to that protein. This greatly hindered the ability of the method to make progress. We therefore used the softmax (non-Gumbel) approximation used by Killoran in the generators employed by FB-GAN for this experiment.

\paragraph{Model selection} One orthogonal issue that we did not focus on in the present work is how best to choose the class and capacity of the generative model, which can be approached using standard methods of model comparison. Herein, we instead simply select one GAN and one VAE, chosen from the literature, because these are simply plug-and-play components of our approach, and not the focus of the present work. In particular, we are not claiming that one of these is better than the other for our problem of input design. It is likely that the usual issues of, for example, blurriness in VAEs and mode collapse in GANs may be quite important~\cite{Engel2018}.

\paragraph{Marginal} The specific procedure used by the Marginal method, given a training set of sequences and corresponding property values is as follows: for each position $i=1, 2,\ldots,L$ the Marginal method splits the training set into four subsets, where the $j^{\text{th}}$ set contains all sequences with sequence element $j$ at a position $i$. It then calculates the average property values corresponding to the sequences in each set, and sets the element at position $i$ to the element whose set has the largest average property value. The resulting sequence is returned. This method is updated slightly for the expression maximization experiments (Section \ref{expmax}) in order to ensure that the constraint that the resulting sequences is translated into the target protein. This is done by marginalizing in the space of codons rather than sequence elements.

\subsection{Maximization of random, noise-free oracles}
\paragraph{Oracle}The networks used in these experiments had architectures $\texttt{Input(L, 4)} \rightarrow \texttt{Flatten(L*4)} \rightarrow \texttt{Dense(50)} \rightarrow \texttt{Dense(50)} \rightarrow \texttt{Dense(1)}$. The weights and biases in these networks were randomly set using the Glorot Uniform scheme \cite{pmlr-v9-glorot10a}.

\subsection{Maximization of a noisy, protein expression oracle}

\paragraph{Oracle Details}  The oracle network used in these experiments had the architecture $\texttt{Input(L, 4)} \rightarrow \texttt{Flatten(L*4)} \rightarrow \texttt{Dense(50)} \rightarrow \texttt{Dense(50)} \rightarrow \texttt{Dense(1)}$. We applied an 85/10/5 train/validation/test split to the sequence/expression pairs from \cite{Cambray2018} and trained the model with the Adam optimizer \cite{DBLP:KingmaB14} and early stopping criteria based on the validation loss (mean squared error). The trained model achieved a Pearson correlation of $0.8$ between true and predicted values on a hold-out test set. A scatter plot of these results is shown in Figure \ref{fig:scatter} below.

\paragraph{Target Details} The experiments here (as well as in the specification case) optimized DNA sequences corresponding to the protein sequence SNILHPLFAVVVVHWSPLKIPSRWKIGVRQYV (the protein with the most experimental measurements). There are 48,693,796,581,408,768 possible sequences that are translated into this protein, of which 418 were tested by the experiments in \cite{Cambray2018} and used as the initial training set for generative models.

\section{Sensitivity to Parameters}

We tested the sensitivity of DbAS results in the first two experiments (Sections \ref{random_nn_section} and \ref{expmax}) to the settings of the $Q$ parameter. By examining example trajectories at different values of $Q$ (Figures \ref{fig:noiseless_q_compare} and \ref{fig:exp_q_compare}), we can see that the results are relatively insensitive the setting of the $Q$ parameter, with only very low values ($<0.4$) performing substantially worse than the settings used in the presented experimental results. Note that $Q$ is only defined for $Q \in [0,1]$), so this encompasses nearly the entire range.

\newpage
\begin{figure}
    \centering
    \includegraphics{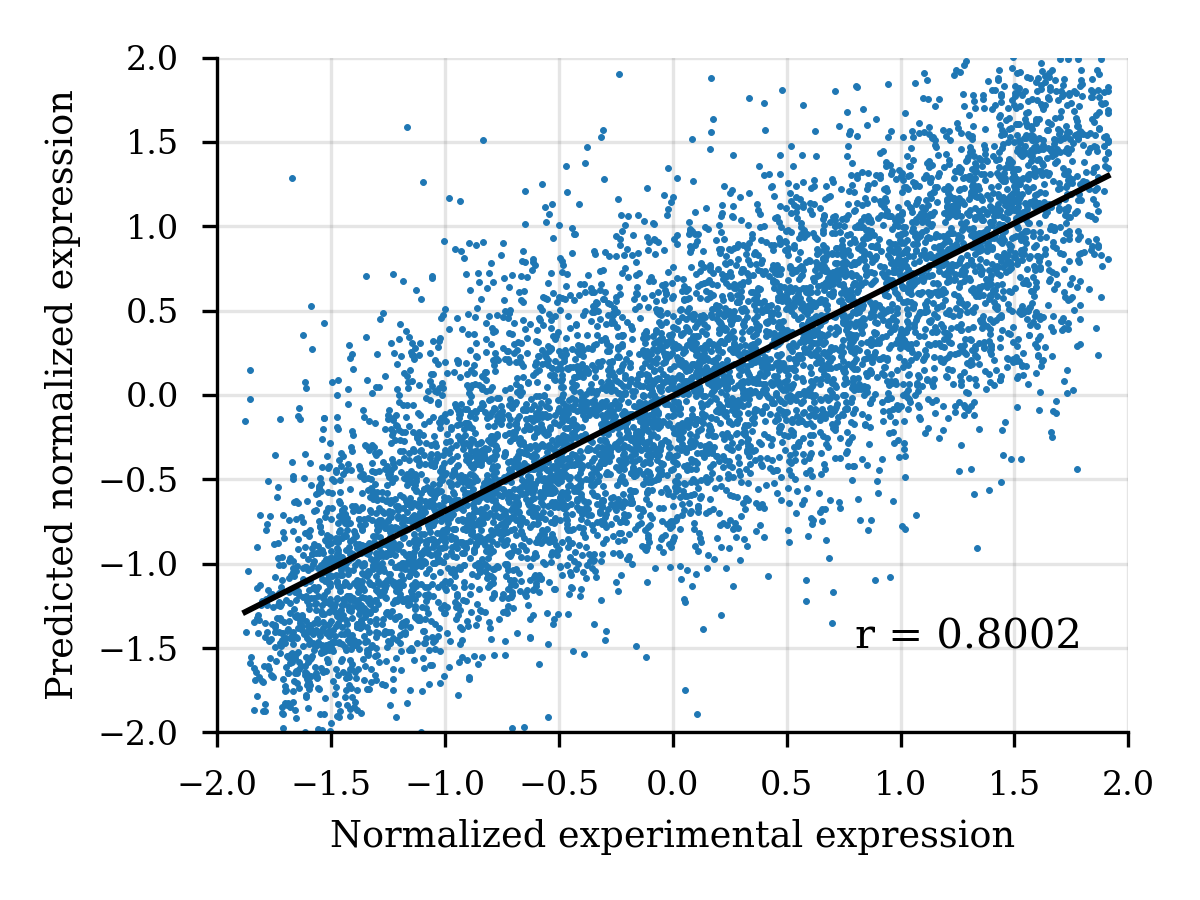}
    \caption{Experimental expression vs. predicted expression on a hold test set. The model used for prediction is that used in all experiments in Sections \ref{expmax} and \ref{expspec}}
    \label{fig:scatter}
\end{figure}

\clearpage
\begin{figure}
    \centering
    \includegraphics{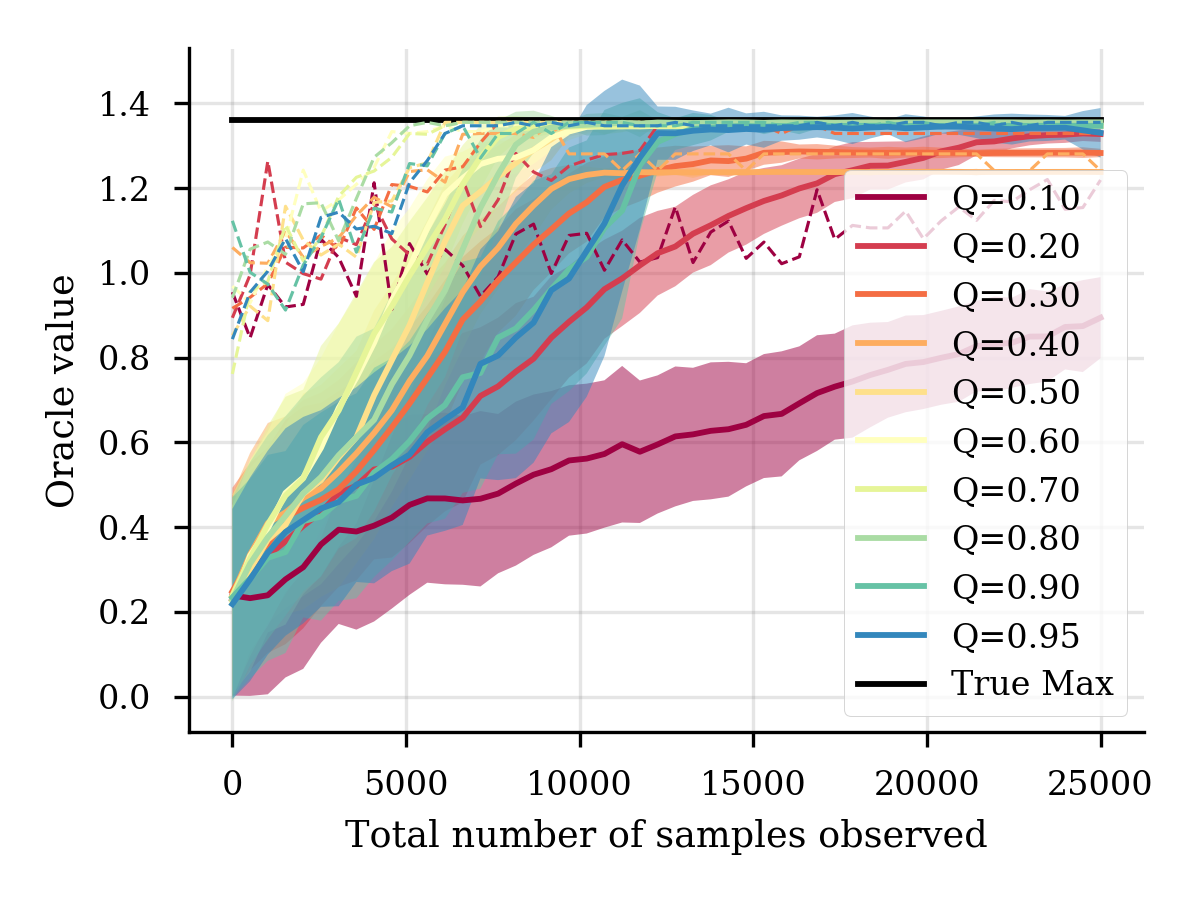}
    \caption{Example trajectories at different settings of the $Q$ parameter in the noiseless random neural network maximization experiment (Section \ref{random_nn_section}) for $L=11$. Note that to highlight the differences between the trajectories, these were given a larger sequence budget than the results presented in Figure \ref{fig:figure1}b.} 
    \label{fig:noiseless_q_compare}
\end{figure}

\begin{figure}
    \centering
    \includegraphics{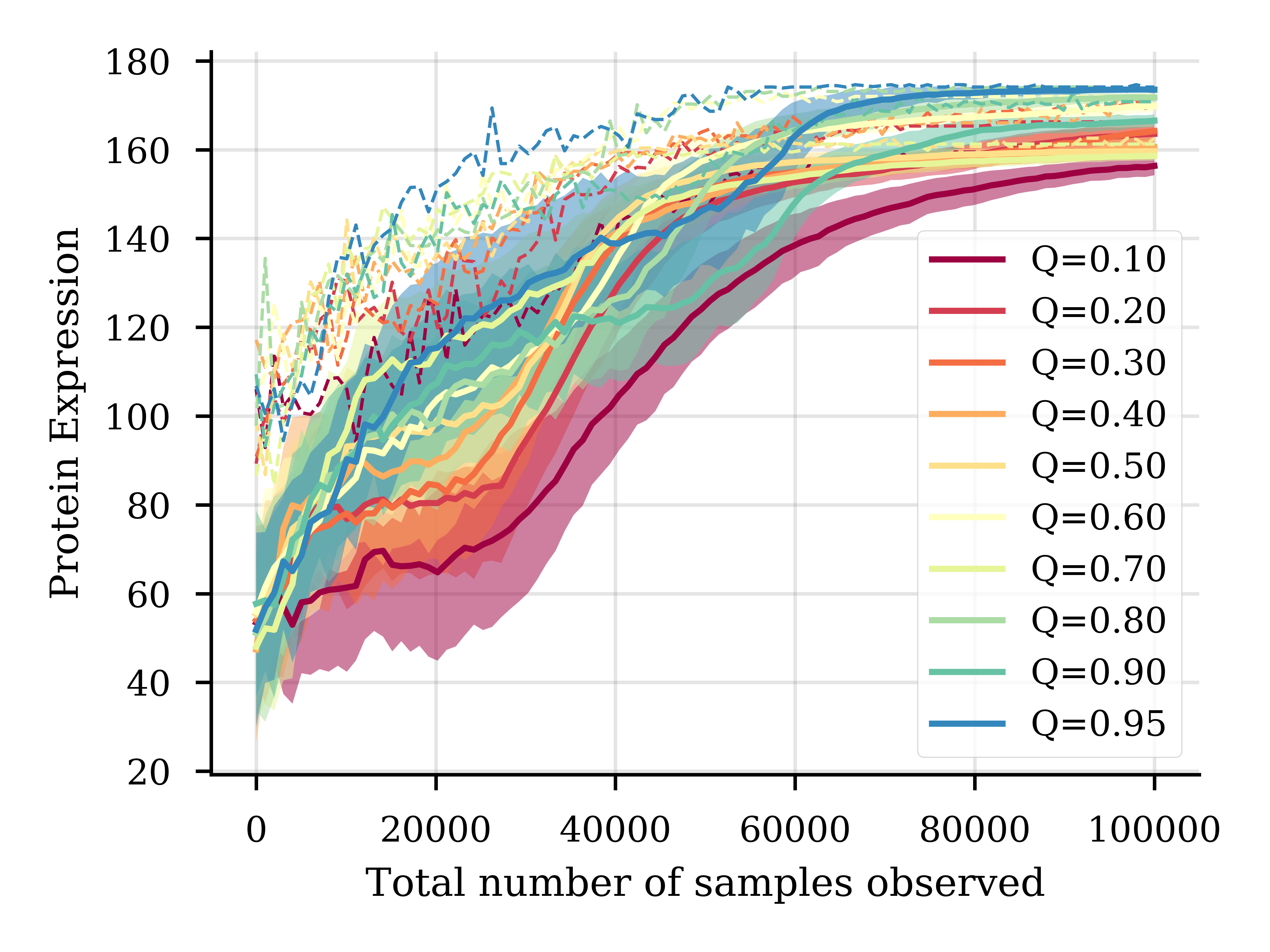}
    \caption{Example trajectories at different settings of the $Q$ parameter in the expression maximization experiment (Section \ref{expmax})}
    \label{fig:exp_q_compare}
\end{figure}


\end{document}